\documentclass[11pt]{article}

\usepackage[final]{acl}

\usepackage{times}
\usepackage{latexsym}
\usepackage[T1]{fontenc}
\usepackage[utf8]{inputenc}
\usepackage{microtype}
\usepackage{inconsolata}

\usepackage{graphicx}
\usepackage{amsmath}
\usepackage{amsfonts}
\usepackage{amssymb}

\usepackage{algorithm}
\usepackage{algorithmic}
\usepackage{booktabs}
\usepackage{array}
\usepackage{stfloats}
\usepackage{balance}
\usepackage{url}
\usepackage{enumitem}
\usepackage[most]{tcolorbox}
\usepackage[skip=3pt]{caption}

\newlist{cotdesc}{description}{1}
\setlist[cotdesc]{
    font=\normalfont\bfseries,
    leftmargin=2.8cm,
    labelsep=0.5cm,
    style=nextline
}

\newtcolorbox{examplebox}{
    enhanced,
    breakable,
    colframe=black!60,
    colback=gray!5,
    boxrule=0.8pt,
    arc=3pt,
    before skip=10pt,
    after skip=10pt,
    left=5pt,
    right=5pt,
    top=5pt,
    bottom=5pt,
    check odd page=true,
    toggle enlargement=none
}

\title{
Superficial Reflection or Genuine Thought?
A Fine-Grained Cognitive Analysis of Large Reasoning Models
}

\author{
\textbf{Yuxiang Chen}\textsuperscript{1,2}\thanks{Equal contribution.} \quad
\textbf{Zuohan Wu}\textsuperscript{3}\footnotemark[1] \quad
\textbf{Ziwei Wang}\textsuperscript{3} \quad
\textbf{Xiangning Yu}\textsuperscript{4} \quad
\textbf{Xujia Li}\textsuperscript{5}\thanks{Corresponding author.} \\
\textbf{Linyi Yang}\textsuperscript{6} \quad
\textbf{Mengyue Yang}\textsuperscript{7} \quad
\textbf{Jun Wang}\textsuperscript{1} \quad
\textbf{Lei Chen}\textsuperscript{3,5} \\
\textsuperscript{1}University College London \qquad
\textsuperscript{2}AI Lab, Yangtze River Delta \qquad
\textsuperscript{3}HKUST (Guangzhou) \\
\textsuperscript{4}Tianjin University \qquad
\textsuperscript{5}HKUST \\
\textsuperscript{6}Southern University of Science and Technology \qquad
\textsuperscript{7}University of Bristol \\
\texttt{leexujia@ust.hk} \\[2pt]
}

\begin{document}
\maketitle

\begin{abstract}
Motivated by the observed human-like behaviours in Large Reasoning Models
(LRMs), this paper introduces a comprehensive taxonomy to characterise atomic
reasoning steps and analyse the reasoning behaviours of LRMs.
Grounded in human cognitive processes, we propose a taxonomy comprising five
groups and seventeen categories. Through this taxonomy, we conduct an
in-depth analysis of contemporary LRMs and distil four actionable takeaways
for model optimisation. Most notably, we reveal that prevailing post-answer
``double-checks'' are largely superficial and rarely yield substantive
revisions. A targeted intervention further shows that explicitly eliciting
richer reflection processes can substantially improve failed self-correction.
To support this large-scale study, we propose CAPO, an automated annotation
method used to construct a dataset of 277,534 reasoning steps with strong
agreement with human expert annotations. We further validate the main
behavioural patterns on a newer reasoning model and a coding domain,
demonstrating the broader applicability of the proposed taxonomy.
All source code and data are available at
\url{https://github.com/hehepig4/psyche}.
\end{abstract}

\section{Introduction}

Large Reasoning Models (LRMs), such as DeepSeek-R1
~\cite{DBLP:journals/corr/abs-2501-12948} and OpenAI
o1~\cite{DBLP:journals/corr/abs-2412-16720}, have demonstrated strong
performance on complex tasks such as mathematics and coding. Unlike
conventional language models, LRMs often produce extended reasoning traces
before giving a final answer, commonly referred to as long
chain-of-thought (CoT)~\cite{DBLP:conf/nips/Wei0SBIXCLZ22}. Interestingly,
these traces frequently contain human-like expressions such as ``wait'',
``let me check'', or ``maybe not'', together with behaviours resembling
recall, hypothesis generation, reconsideration, and self-checking. Such
observations have motivated increasing interest in understanding LRM
reasoning from perspectives inspired by human cognition.

Existing analyses, however, remain relatively coarse-grained. A common
perspective interprets LRM reasoning through high-level cognitive analogies
such as System~1 and System~2 reasoning~\cite{wang2025tutorial}. While useful
for conceptualising extended reasoning, such distinctions offer limited
resolution for characterising the diverse behaviours within a long CoT.
For example, ``reflection'' may refer to merely checking a previous result,
identifying the cause of an error, considering alternatives, or actually
changing the reasoning strategy. Treating these processes as a single
behaviour can obscure important differences between successful and
unsuccessful reasoning.

Meanwhile, CoT outputs should not be interpreted as direct simulations of
human neural or cognitive mechanisms. Human cognition emerges from
biologically grounded processes, whereas LRMs acquire their behaviours
through large-scale data-driven training and reinforcement learning
~\cite{DBLP:journals/corr/abs-2501-12948,wang2024openr,feng2023alphazero}.
We therefore use human cognitive processes as \emph{descriptive templates}
for observable reasoning behaviours rather than claims about the internal
mechanisms of LRMs.

\begin{figure*}[t]
\centering
\includegraphics[width=1\linewidth]{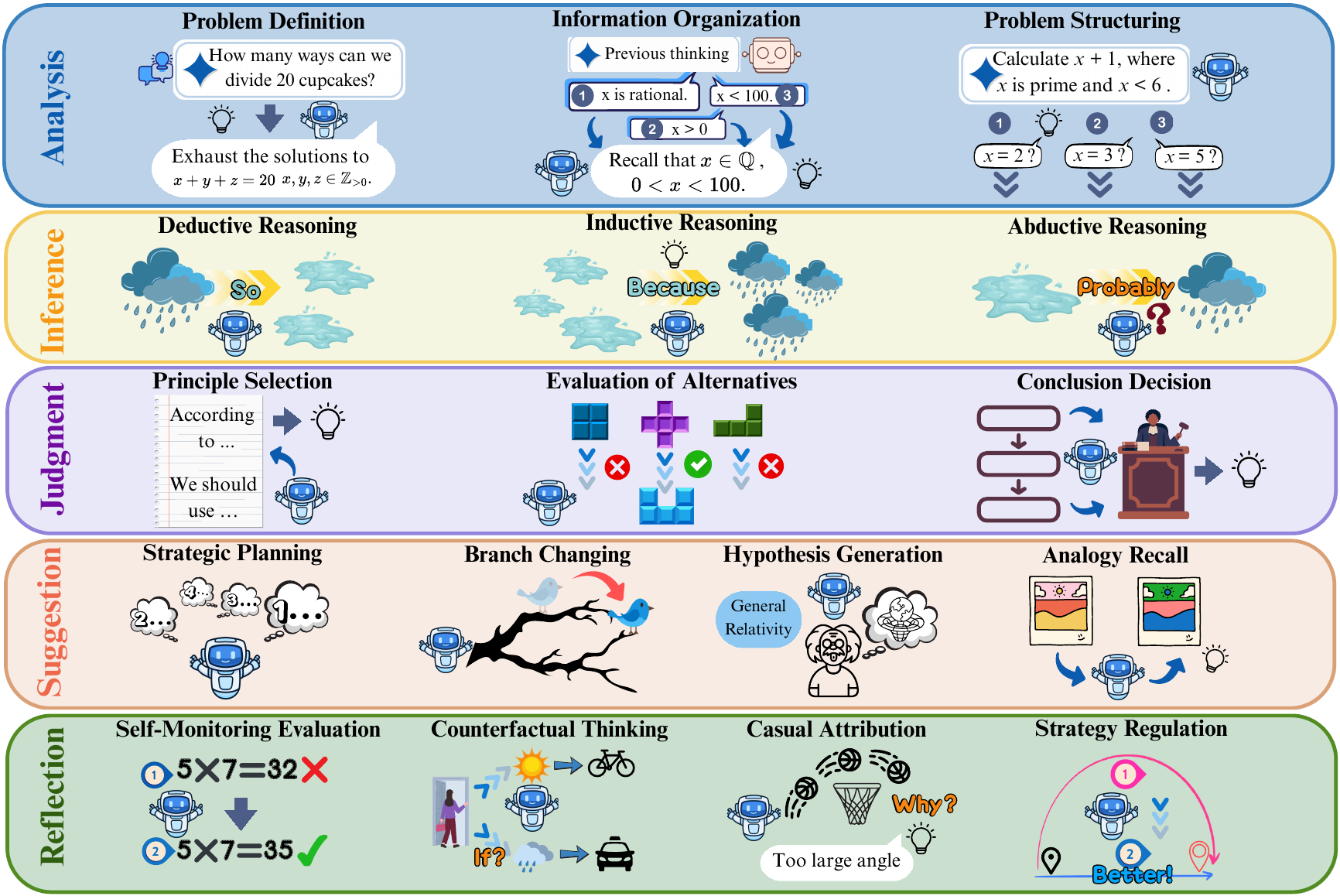}
\caption{A brief illustration of the proposed taxonomy from human cognitive perspectives.}
\label{fig:Taxonomy}
\vspace{-1em}
\end{figure*}

Motivated by this perspective, we introduce a fine-grained taxonomy for
analysing individual steps in LRM reasoning traces. Drawing inspiration from
formal logic~\cite{gentzen1935,prawitz1965}, education
theory~\cite{vygotsky1978mind,wood1976role,collins1989cognitive}, and
cognitive psychology
~\cite{flavell1979metacognition,chi1989self,dunlosky2013learning}, the
taxonomy contains \textbf{five high-level groups and seventeen fine-grained
categories}. This hierarchical design preserves commonality at the group
level while distinguishing processes that may play very different roles
during reasoning, such as self-monitoring, causal attribution, hypothesis
generation, and information organisation.

The fine granularity of this taxonomy makes large-scale manual annotation
prohibitively expensive. We therefore propose \textbf{Constrained Automatic
Prompt Optimisation (CAPO)}, which aligns an LLM annotator with expert labels
through iterative prompt optimisation. CAPO achieves a micro Cohen's
$\kappa$ of \textbf{0.811} and a macro $\kappa$ of \textbf{0.594} against
expert annotations, while the principal behavioural signals underlying our
analysis are also preserved on the human-annotated subset. Using this
framework, we construct a corpus of \textbf{277,534 annotated mathematical
reasoning steps}, including \textbf{9,841 human annotations} and
\textbf{267,693 CAPO annotations}.

Our analysis reveals four recurring patterns in current LRMs:
\begin{enumerate}[leftmargin=*, nosep]

\item \textbf{Information organisation.}
Successful CoTs more frequently recall and reorganise important intermediate
information, helping maintain relevant context over long reasoning
trajectories.

\item \textbf{Analogy \& hypothesis.}
Hypothesis generation and analogy recall tend to occur later and with weaker
justification in unsuccessful reasoning trajectories.

\item \textbf{Reflection.}
Native post-answer ``double-checks'' are predominantly shallow
self-monitoring and rarely produce substantive revisions. In a targeted
intervention, explicitly eliciting richer reflection corrects
\textbf{43 of 62 (69.3\%)} previously failed self-correction cases.

\item \textbf{Redundancy.}
Long reasoning traces contain substantial numbers of steps with low estimated
necessity, suggesting that longer reasoning does not necessarily imply more
effective reasoning.

\end{enumerate}

We further examine the framework on the newer MiniMax-M2 model and on
CodeForces reasoning traces. Several core behavioural patterns transfer
across model generations and reasoning domains, while the taxonomy also
captures meaningful changes in newer LRMs. Detailed robustness analyses are
reported in Appendix~\ref{sec:add_exp}.

In summary, our contributions are:
\begin{enumerate}[leftmargin=*, nosep]

\item We introduce a five-group, seventeen-category taxonomy that provides a
fine-grained descriptive framework for analysing LRM reasoning trajectories.

\item We propose CAPO for scalable step-level annotation and validate its
agreement with expert annotations.

\item We construct a large annotated reasoning corpus and identify four
empirical patterns concerning information organisation, speculative
reasoning, reflection, and redundancy, complemented by targeted intervention
and cross-model and cross-domain validation.

\end{enumerate}

\vspace{-5pt}
\section{Related Work}\label{sec:related}

With the remarkable success of reasoning models such as OpenAI's o1, research has intensified in this area. Following \cite{DBLP:journals/corr/abs-2503-09567}, two key behaviour groups are often distinguished.
\textbf{Deep reasoning}, covering step-wise inference and planning, has been strengthened via prompt engineering (e.g., chain-of-thought) \cite{DBLP:conf/nips/Wei0SBIXCLZ22}, specialised decoding/control structures \cite{DBLP:conf/aaai/BestaBKGPGGLNNH24,DBLP:conf/nips/YaoYZS00N23}, and targeted training \cite{DBLP:conf/nips/ZelikmanWMG22}.
\textbf{Reflection} constitutes a second core behaviour group \cite{DBLP:journals/corr/abs-2503-09567}, where methods introduce feedback-then-refinement cycles to revise drafts or plans \cite{DBLP:journals/corr/abs-2502-11520,DBLP:conf/acl/WangLSXDLCWS24,DBLP:conf/iclr/KumarZASCSBIBRZ25}.
Beyond ``how to do'' reasoning, recent studies probe where improvement opportunities lie, examining factors such as problem complexity \cite{DBLP:journals/corr/abs-2506-06941}, inference-time scaling \cite{DBLP:journals/corr/abs-2410-13639}, error types \cite{DBLP:journals/corr/abs-2502-19361}, and confidence \cite{DBLP:journals/corr/abs-2505-14489}.
However, existing definitions and taxonomies of CoTs remain limited in coverage. A broader, step-wise perspective in an interdisciplinary human lens remains underexplored.

To date, one study in this line first \cite{DBLP:journals/corr/abs-2504-07128} proposes a rudimentary four-category taxonomy, reporting preliminary findings as a minor component of a larger agenda. Substantially richer structure nonetheless appears attainable. The idea of classifying thought has a long lineage: from Plato’s \emph{Statesman} on classification in governance and knowledge \cite{plato_statesman} and Aristotle’s \emph{Categories} on distinctions underlying reasoning \cite{aristotle_categories_edghill}, to modern accounts in logic (proof theory and natural deduction) \cite{gentzen1935,prawitz1965}, education theory (scaffolding and cognitive apprenticeship) \cite{vygotsky1978mind,wood1976role,collins1989cognitive}, and cognitive psychology (metacognition and self-explanation) \cite{flavell1979metacognition,chi1989self,dunlosky2013learning}.
Building on these traditions, this paper advances a fine-grained, step-wise CoT taxonomy and a large-scale annotation methodology (CAPO), enabling comprehensive behavioural analysis beyond coarse stage-based schemes.

\section{A CoT Taxonomy through a Human Cognitive Lens}
\label{sec:Human-Thought-Taxonomy}

We develop a principled taxonomy for analysing the steps of CoT reasoning
produced by LRMs. Recall that CoT sequences are not direct simulations of
neural or cognitive mechanisms. Thus, we ground our taxonomy in the
pedagogical tradition, treating human thought patterns as expert templates
that serve as descriptive lenses for classifying the expression of reasoning
in LLMs. Subsequently, we draw inspiration from multiple disciplines that
have long examined the nature and structure of human reasoning. As depicted
in Figure~\ref{fig:Taxonomy}, we refine five high-level categories into
seventeen finer-grained subcategories. This hierarchical structure facilitates
clear distinctions between reasoning types while retaining their commonality
at the higher level. For concrete illustrations of each subcategory, please
refer to Appendix~\ref{sec:appendix_cot}.

\paragraph{Operationalising reasoning steps.}
We operationalise a reasoning step using the natural paragraph boundaries
(``\texttt{\textbackslash n\textbackslash n}'') produced by the LRMs studied
in this work. These boundaries typically correspond to coherent ``thought
blocks'' in long CoTs. Empirically, \textbf{82.2\%} of the resulting segments
receive exactly one mental-process label, while only a small fraction contain
more than two labels. We further conduct a robustness analysis on 100 randomly
sampled CoTs using sentence-level segmentation. The principal behavioural
patterns remain broadly consistent, while sentence-level splitting increases
the number of annotation units by approximately \textbf{1.63$\times$}.
These observations suggest that paragraph-level segmentation provides a
practical balance between semantic coherence, analytical granularity, and
annotation cost.

\vspace{2pt}
\textit{\textbf{Analysis}.}
Analysis entails decomposing complex ideas into constituent parts to understand
their interrelations. This phase corresponds to model behaviours where abstract
tasks are broken down into subtasks, either explicitly through prompting or
implicitly via attention~\cite{mayer1998cognitive}. From a process-level
perspective, this involves:

\begin{itemize}[leftmargin=*, nosep]
\item \textbf{Problem Definition (A.PD)}:
Clarifying the problem's core challenge by paraphrasing the question,
identifying its type, or surfacing hidden goals and constraints.

\item \textbf{Problem Structuring (A.PS)}:
Breaking the problem into logical parts or subgoals, often by outlining a
solution plan or isolating knowns and unknowns.

\item \textbf{Information Organisation (A.IO)}:
Reviewing or restructuring prior information, such as results or premises,
to support upcoming reasoning~\cite{punia2023logical}.
\end{itemize}

In LLMs, analysis functions as a stabilising scaffold, enabling the model to
reason with awareness of context, dependencies, and previously computed
results~\cite{punia2023logical}. Although such steps may not directly advance
the final answer, they can play a metacognitive role in maintaining coherence
and preserving relevant information throughout long-form reasoning.

\vspace{2pt}
\textit{\textbf{Inference}.}
From formal logic and philosophy of science, we incorporate canonical
paradigms of inference, including deduction, induction, and
abduction~\cite{carnap1971studies,johnsonlaird1991deduction}. These forms of
inference constitute the backbone of reasoning processes and correspond to
different ways in which assertions are drawn from premises, whether
necessarily (deduction), probabilistically (induction), or plausibly
(abduction):

\begin{itemize}[leftmargin=*, nosep]
\item \textbf{Deductive Reasoning (I.DR)}:
Applying general rules to derive logically certain conclusions. This is common
in tasks such as mathematical proofs, where valid premises guarantee correct
outcomes~\cite{carnap1971studies}.

\item \textbf{Inductive Reasoning (I.IR)}:
Generalising patterns from specific examples. This is typical in empirical
reasoning or analogical settings, where conclusions are probable but not
certain~\cite{orourke1997automated}.

\item \textbf{Abductive Reasoning (I.AR)}:
Inferring the most plausible explanation for an observation. Although
uncertain, it is important for hypothesis generation and commonsense
reasoning~\cite{johnsonlaird1991deduction}.
\end{itemize}

For instance, deductive structures can be observed in arithmetic CoT tasks,
inductive patterns emerge in classification or analogy tasks, and abductive
moves often appear when the model speculates about hidden causes or
intentions. As emphasised in~\cite{dewey1910how}, inference operates as the
connective tissue between intermediate reasoning steps, enabling a model to
construct coherent lines of argumentation or problem-solving trajectories.

\vspace{2pt}
\textit{\textbf{Judgment}.}
Judgment involves comparing alternatives and selecting solutions based on
principled reasoning, aligning with pedagogical models of critical thinking,
argument evaluation, and decision-making. Within human reasoning theory,
judgment typically involves weighing competing hypotheses, assessing
consistency with prior knowledge, and selecting the most justified course of
action. In LLMs, judgment refers to the evaluative process by which an agent
compares alternative solution paths and determines which is most appropriate
based on prior reasoning, including three subtypes:

\begin{itemize}[leftmargin=*, nosep]
\item \textbf{Principle Selection (J.PS)}:
Choosing appropriate logical principles, ethical rules, or task-specific
criteria to guide decision-making.

\item \textbf{Evaluation of Alternatives (J.EA)}:
Comparing competing reasoning paths or hypotheses to select the most viable
direction.

\item \textbf{Conclusion Decision (J.CD)}:
Making a final commitment to an answer or solution, justified by prior
reasoning steps.
\end{itemize}

For disambiguation with the Suggestion type introduced next, judgments are
conclusions drawn from the preceding reasoning process, whereas suggestions
introduce potential directions or information before such evaluation.

\vspace{2pt}
\textit{\textbf{Suggestion}.}
Suggestion is informed by studies on spontaneous idea generation, creativity,
and the psychological phenomenon of suggestion
itself~\cite{gheorghiu1987suggestion}, which highlight how new directions in
thought can emerge without immediate justification.

In our taxonomy, suggestion refers to the generative act of proposing new
ideas that extend beyond the direct content of the problem. It encompasses
heuristic and forward-looking reasoning behaviours that introduce novel
directions, potential solution paths, or speculative constructs before any
evaluation takes place.

\begin{itemize}[leftmargin=*, nosep]
\item \textbf{Strategic Planning (S.SP)}:
Proposing a plan or high-level roadmap for how the problem might be
approached. This often appears as a declarative intention to structure
upcoming reasoning steps (e.g., ``First, I will try a substitution, then
check for symmetry'').

\item \textbf{Branch Changing (S.BC)}:
Initiating a shift from the current reasoning path to an alternative one,
typically when the existing direction is perceived as unproductive.

\item \textbf{Hypothesis Generation (S.HG)}:
Formulating a tentative explanation or educated guess based on limited
evidence. It is relevant to tasks involving hidden rules, causality, or
implicit goals, where the next step is unclear.

\item \textbf{Analogy Recall (S.AR)}:
Recalling a past experience, familiar structure, or well-known problem to
inform the current task. Analogical suggestion often acts as a conceptual
scaffold, helping the model bridge from known solutions to new domains.
\end{itemize}

Understanding and identifying Suggestion steps within CoT sequences thus
provides insight into how models initiate, branch, and explore within complex
problem spaces.

\vspace{2pt}
\textit{\textbf{Reflection}.}
Reflection provides a cognitive model for meta-level reasoning and error
awareness~\cite{dewey1910how}. It represents a metacognitive capability to
step outside the current stream of reasoning and critically evaluate its
validity, necessity, and efficiency:

\begin{itemize}[leftmargin=*, nosep]
\item \textbf{Self-Monitoring Evaluation (R.SME)}:
Reviewing the reasoning process so far and checking for errors or
inconsistencies in logic.

\item \textbf{Counterfactual Thinking (R.CT)}:
Considering alternative actions or decisions and reasoning about what might
have occurred under different conditions. This can be used to reassess current
reasoning or outcomes through ``what-if'' scenarios.

\item \textbf{Causal Attribution (R.CA)}:
Analysing the reasons behind success or failure by identifying the factors or
decisions that contributed to the result.

\item \textbf{Strategy Regulation (R.SR)}:
Adjusting the current overall reasoning or problem-solving strategy based on
feedback or prior reflection.
\end{itemize}

Notably, existing studies such as~\cite{DBLP:journals/corr/abs-2410-18982}
have attributed reflection as an important capability underlying successful
reasoning in LRMs. Reflection allows backward examination of generated steps,
identification of potential errors, and reconsideration of whether the current
direction remains appropriate. As we show later, however, different forms of
reflection exhibit substantially different behavioural patterns, motivating
the need to distinguish self-monitoring from deeper processes such as causal
attribution and strategy regulation.

\paragraph{Why fine-grained categories?}
The seventeen subcategories are intended as a descriptive analytical
resolution rather than a claim that they form the unique decomposition of
machine reasoning. To examine whether this finer granularity provides
additional empirical information, we also aggregate the labels into the five
parent categories and repeat our analysis. The coarser representation
substantially obscures category-specific patterns: for example, the aggregate
Suggestion category exhibits no significant overall difference between
successful and unsuccessful reasoning, despite distinct behaviours of
Hypothesis Generation and Analogy Recall revealed by the fine-grained
analysis. This suggests that the finer taxonomy preserves behavioural signals
that can be diluted when heterogeneous processes are merged.

\section{CAPO: Scalable LLM-Assisted Annotation}
\label{sec:category_assignment}

Given the prohibitive cost of manual annotation---for example, 931 AIME CoTs
contain 177,687 reasoning steps---and the demonstrated capability of LLMs for
data labelling~\cite{DBLP:conf/acl/ChiangL23,DBLP:journals/corr/abs-2501-10970},
we leverage LLMs to automate the assignment of our taxonomy categories.

The major challenge lies in aligning LLM annotations with human expert
standards~\cite{DBLP:journals/corr/abs-2501-10970}. Conventional techniques
such as \textit{fine-tuning} (FT) and \textit{in-context learning} (ICL) are
less suitable for our setting. FT requires relatively large amounts of
labelled data, whereas ICL becomes expensive for long reasoning trajectories.
Since classifying an individual step may depend on its preceding reasoning
context, incorporating multiple full-context demonstrations can substantially
increase prompt length and exacerbate long-context limitations, including the
\textit{lost-in-the-middle} phenomenon~\cite{DBLP:journals/tacl/LiuLHPBPL24}.
To address these issues, we propose \textbf{Constrained Automatic Prompt
Optimisation (CAPO)}.

Inspired by iterative human learning, CAPO optimises the annotation prompt by
analysing discrepancies between zero-shot predictions and expert annotations.
We formulate this process within a genetic algorithm (GA) framework comprising
three operators:
(1) \textbf{Mutation}, where the LLM summarises annotation ``tips'' from
individual alignment errors and incorporates them into the prompt;
(2) \textbf{Reproduction}, where a meta-process synthesises an improved prompt
by combining two high-performing prompt variants; and
(3) \textbf{Elimination}, where candidate prompts are evaluated and
underperforming variants are discarded to enable iterative improvement.

To mitigate overfitting to the small expert-labelled training set, we further
introduce a \textit{tripartite prompting strategy} that differs from existing
GA-based prompt optimisation methods~\cite{evoprompt,gaapo,strago}.
Specifically, we partition the prompt into three regions: the
\textbf{constant region}, which contains invariant task formats; the
\textbf{variable region}, which contains taxonomy descriptions with only
constrained modifications; and the \textbf{mutable region}, which remains
open-ended and stores task-specific annotation guidance. This design preserves
the core taxonomy definitions while allowing CAPO to learn flexible annotation
strategies from observed human--model discrepancies. Representative annotation
tips learned by CAPO are provided in Appendix~\ref{apx:capo_details}.

\subsection{CAPO Evaluation}
\label{sec:exp_apo}

We evaluate CAPO using \textit{Gemini-2.5-Flash} against a
Retrieval-Augmented Generation (RAG) baseline
~\cite{DBLP:journals/corr/abs-2504-07128}, which performs ICL with retrieved
expert-labelled exemplars. The evaluation is conducted on a split of our
human-annotated AIME$^\star$ and HMMT data.

\begin{figure}[!tbp]
    \centering
    \includegraphics[width=1\linewidth]{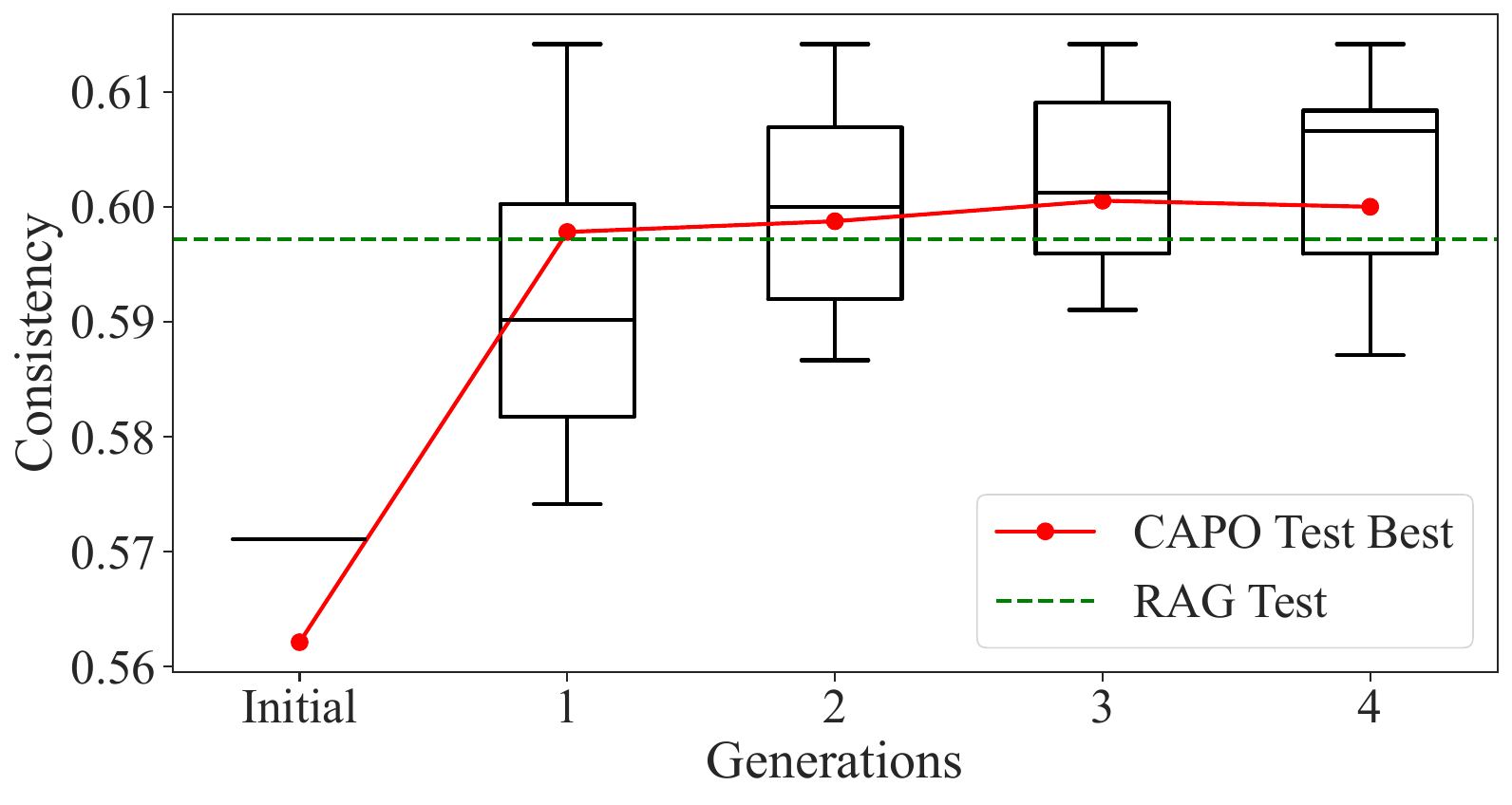}
    \caption{Annotation consistency of CAPO and the RAG baseline.
    The boxes show the consistency distribution of candidate prompts on the
    training population, while the red line denotes the test consistency of
    the best validation-selected prompt.}
    \label{fig:consistencies}
    \vspace{-1em}
\end{figure}

As shown in Figure~\ref{fig:consistencies}, CAPO surpasses the RAG baseline
after a single optimisation round while avoiding the substantially longer
contexts required by retrieved ICL demonstrations. These results show that
prompt optimisation provides an effective and comparatively context-efficient
way to scale our taxonomy annotation.

\paragraph{Agreement with expert annotations.}
Because CAPO provides the majority of annotations used in our subsequent
analysis, we additionally quantify human--machine agreement using Cohen's
$\kappa$. CAPO achieves a \textbf{micro $\kappa$ of 0.811} and a
\textbf{macro $\kappa$ of 0.594}. The lower macro agreement is mainly
concentrated in rare categories, for which substantially fewer expert-labelled
examples are available.

More importantly, we separately examine whether the behavioural signals used
in our main findings are preserved between the human- and CAPO-annotated
subsets. Table~\ref{tab:human_machine_consistency} reports the four categories
most directly involved in the subsequent takeaways. In each case, the
machine-estimated mean difference falls within the corresponding interval
estimated from human annotations.

\begin{table}[!tbp]
\centering
\resizebox{0.90\columnwidth}{!}{\begin{tabular}{c|c|c|c}
\toprule
\textbf{Process} & \textbf{Human interval} &
\textbf{Machine MD} & \textbf{Within interval} \\
\midrule
A.IO  & $[-0.067,\,+0.094]$ & $+0.011$ & \checkmark \\
S.HG  & $[-0.048,\,+0.018]$ & $-0.041$ & \checkmark \\
S.AR  & $[-0.061,\,+0.008]$ & $-0.030$ & \checkmark \\
R.SME & $[-0.073,\,+0.092]$ & $-0.018$ & \checkmark \\
\bottomrule
\end{tabular}}
\caption{Comparison of the principal behavioural effects estimated from
human and CAPO annotations. MD denotes the mean difference between correct
and incorrect CoTs. Full category-level results are provided in the appendix.}
\label{tab:human_machine_consistency}
\vspace{-1em}
\end{table}

Therefore, although CAPO does not reproduce every expert annotation perfectly,
the principal signals underlying our subsequent analysis are also supported by
the expert-labelled subset. Agreement is lower for several rare reflection
categories, such as R.CT and R.CA; we discuss this limitation and provide the
full category-level comparison in Appendix~\ref{apx:capo_details}.

\section{Mental-Process Patterns in LRM Reasoning}
\label{sec:exp_takeaways}

With the taxonomy and annotation framework established, we now analyse
real-world LRM reasoning trajectories to identify mental-process patterns
associated with successful and unsuccessful reasoning.

\vspace{1pt}
\textit{\textbf{Dataset}.}
Our core analysis focuses on mathematical problem solving. For human
annotation, we collect thirty problems from the 2025
AIME$^\star$~\cite{AIME2025} and HMMT~\cite{balunovic_srimatharena_2025},
covering major areas of high-school mathematics. To scale the analysis, we
further include MATH~\cite{DBLP:conf/nips/HendrycksBKABTS21} and
AIME~\cite{di_zhang_2025}, which are annotated using CAPO as introduced in
Section~\ref{sec:category_assignment}. Unless otherwise specified, the
reasoning CoTs are generated by DeepSeek-R1
(0120)~\cite{DBLP:journals/corr/abs-2501-12948}, a representative open-source
reasoning model with 671B parameters.

The core mathematical dataset contains \textbf{277,534 annotated reasoning
steps}, including \textbf{9,841 expert annotations} and
\textbf{267,693 CAPO annotations}. We additionally collect a commonsense
subset for qualitative analysis. Dataset statistics are summarised in
Table~\ref{tab:num_annotations}, and links are provided in
Appendix~\ref{apx:data}.

For the commonsense subset, DeepSeek-R1 answers almost all questions correctly.
Because the analyses below explicitly compare correct and incorrect CoTs, we
exclude this subset from the corresponding correctness-based analyses, as
discussed in Appendix~\ref{sec:discussion}.

\vspace{1pt}
\textit{\textbf{Annotation protocol}.}
Following practices in~\cite{DBLP:journals/corr/abs-2502-19361}, we segment
long CoTs using the natural paragraph delimiter
`\textbackslash{}n\textbackslash{}n'. Annotators identify \textbf{all}
applicable process tags for \textbf{each} reasoning step, making annotation a
multi-label classification task. Human annotators receive training and
discussion sessions based on the taxonomy in
Section~\ref{sec:Human-Thought-Taxonomy}; large-scale annotations are produced
using CAPO. We further examine the robustness of the segmentation choice in
Section~\ref{sec:add_exp}.

\begin{table}[!tbp]
  \centering
  \resizebox{0.95\columnwidth}{!}{\begin{tabular}{c|c|c|c}
      \toprule
      \textbf{Name} & \textbf{Correct} & \textbf{Incorrect} & \textbf{Steps} \\
      \midrule
      MATH & 963 (96.3\%) & 37 (3.7\%) & 90,006\\
      AIME & 810 (87.0\%) & 121 (13.0\%) & 177,687\\
      \midrule
      ComS & \multicolumn{2}{c|}{1000} & 7,710 \\
      \midrule
      HMMT & 9 (64.3\%) & 5 (35.7\%) & 4,375\\
      AIME$^\star$ & 6 (37.5\%) & 10 (62.5\%) & 5,466\\
      \midrule
      ComS$^\star$ & \multicolumn{2}{c|}{30} & 254\\
      \bottomrule
    \end{tabular}}
  \caption{Number of annotated CoTs and reasoning steps. The top three rows
  use CAPO annotations, whereas the bottom three use human annotations. The
  core mathematical analysis excludes the two ComS rows.}
  \label{tab:num_annotations}
  \vspace{-1em}
\end{table}

\begin{figure}[!tbp]
 \centering
 \includegraphics[width=1\linewidth]{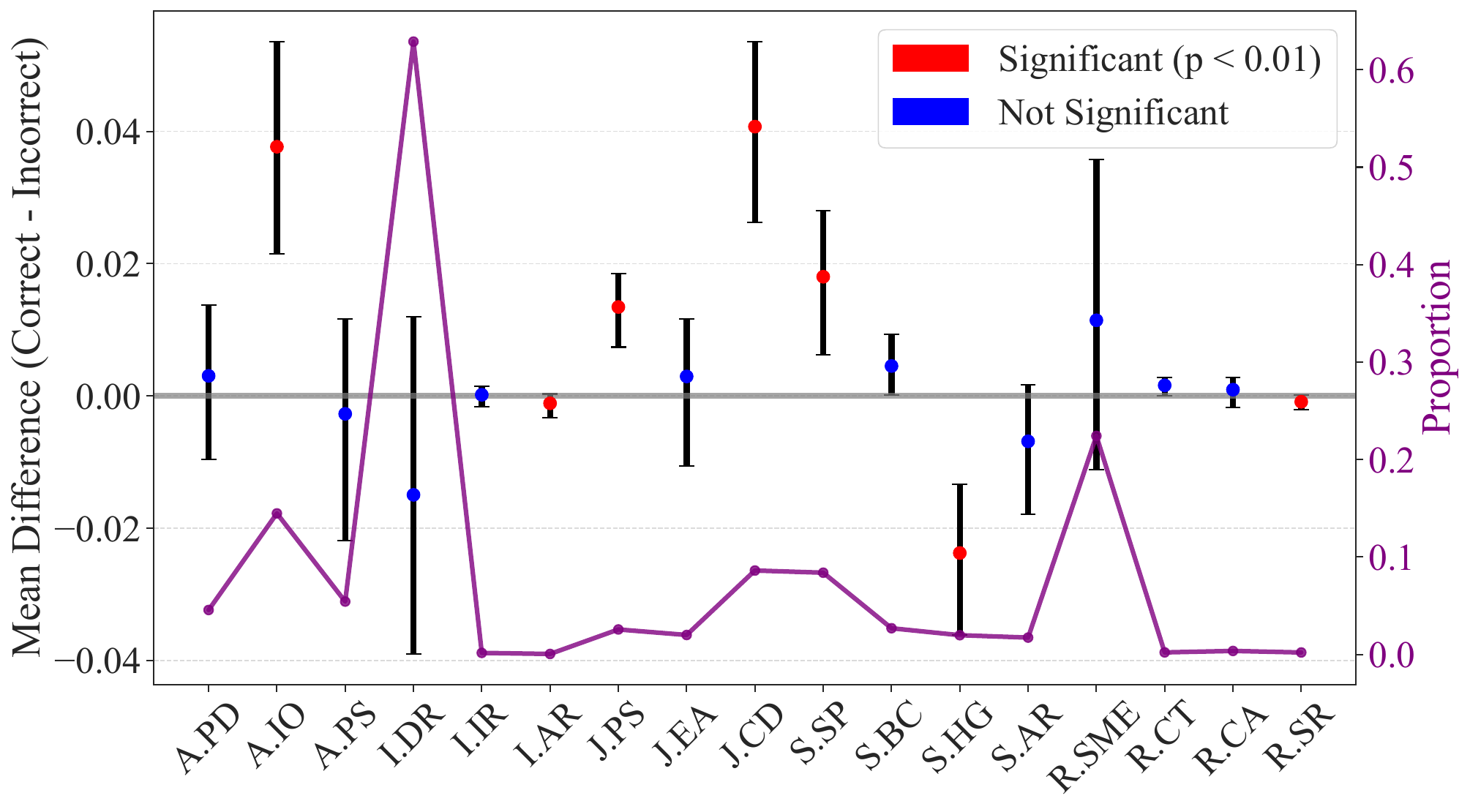}
 \caption{Proportional differences between correct and incorrect CoTs.
 Red dots indicate statistically significant differences. The purple line
 denotes the overall proportion of each mental process.}
 \label{fig:mean_diff}
 \vspace{-1em}
\end{figure}

Based on these trajectories, we investigate which mental-process patterns are
associated with successful reasoning. We compress each CoT into a
17-dimensional feature vector in $[0,1]^{17}$, where each dimension denotes
the proportion of one mental process. For each category, we test whether its
proportion differs significantly between correct and incorrect CoTs.
Figure~\ref{fig:mean_diff} summarises the results.

\vspace{1pt}
\textit{\textbf{Recalling forgotten}.}
LRMs frequently revisit previously established information through
Analysis.Information Organisation (A.IO). In long reasoning trajectories,
periodically restating intermediate milestones may help keep important
information available to subsequent reasoning and mitigate the well-known
\textit{lost-in-the-middle} phenomenon
~\cite{DBLP:journals/tacl/LiuLHPBPL24}.

Figure~\ref{fig:mean_diff} shows that the proportion of A.IO is more than
$0.04$ lower in incorrect CoTs than in correct ones. The following example
illustrates a failure in which the model states two relevant conditions but
subsequently loses track of the second:

\begin{tcolorbox}[colframe=black]\small
\begin{enumerate}[leftmargin=*, nosep]
 \item[\textbf{52:}] \textit{To have G in the last word, two conditions hold:}
 \item[\textbf{53:}] \textit{1. G is a first letter.}
 \item[\textbf{54:}] \textit{2. All other first letters are less than G.}
 \item[\textbf{...}] \textbf{Step 53 discussed, step 54 forgotten}.
\end{enumerate}
\end{tcolorbox}

\begin{tcolorbox}[
title=\textbf{Takeaway 1},
colframe=white!90!black,
coltitle=black,
halign title=center,
outer arc=1mm]
Information organisation is strongly associated with successful reasoning.
This motivates training or post-training strategies that encourage models to
preserve, recall, and reorganise important intermediate information throughout
long reasoning trajectories.
\end{tcolorbox}

\vspace{1pt}
\textit{\textbf{Specious intuitions}.}
Suggestion.Hypothesis Generation (S.HG) and Suggestion.Analogy Recall (S.AR)
introduce candidate ideas that are not necessarily accompanied by immediate
rigorous derivation. As shown in Figure~\ref{fig:mean_diff}, these processes
occur more frequently in failed reasoning trajectories. One possible
interpretation is that models become more likely to invoke speculative
directions after earlier attempts fail.

We further examine \emph{when} these processes occur.
Table~\ref{tab:shg_sar} reports their average relative positions, where each
step position is normalised to $[0,1]$ by CoT length. Both S.HG and S.AR occur
significantly later in incorrect trajectories. This does not imply that analogy
or hypothesis generation is inherently harmful; rather, \emph{late and
insufficiently justified} use of these processes is associated with
unsuccessful reasoning.

A representative failure case is shown below:

\begin{tcolorbox}[coltitle=black]\small
\begin{enumerate}[leftmargin=*, nosep]
 \item[\textbf{244}:] \textbf{Suggestion.Analogy Recall}:
 \textit{However, \textbf{according to references I found}, the number of
 $3\times3$ Sudoku grids (which is a different problem) is
 $9! \times 72 \times 1$. However, for the $3\times9$ grid, the count might
 indeed be $\frac{9!\times6!\times3!}{(3!)^3}$.
 \textbf{Let's accept this for now.}}
 \item[\textbf{252}:] \textbf{Judgment.Conclusion Decision}:
 \textit{\textbf{Given the time I've spent and the references I recall},
 I think the answer is $47$.} (\textbf{wrong})
\end{enumerate}
\end{tcolorbox}

\begin{table}[!tbp]
 \centering
 \begin{tabular}{cccc}
 \toprule
 \textbf{Name} & \textbf{Pos. co.} & \textbf{Pos. inco.} & \textbf{P-value} \\
 \midrule
 S.HG & 0.35 & 0.47 (+0.12) & $4.2\times10^{-7}$\\
 S.AR & 0.40 & 0.48 (+0.08) & $6.3\times10^{-3}$\\
 \bottomrule
 \end{tabular}
 \caption{Average relative positions of Hypothesis Generation and Analogy
 Recall. Both occur significantly later in incorrect CoTs ($p<0.01$).}
 \label{tab:shg_sar}
\end{table}

\begin{tcolorbox}[
title=\textbf{Takeaway 2},
colframe=white!90!black,
coltitle=black,
halign title=center,
outer arc=1mm]
LRMs exhibit preliminary abilities of analogy recall and hypothesis generation,
but these behaviours tend to appear later in unsuccessful reasoning and may be
invoked without sufficient justification. This motivates future training
objectives that distinguish productive exploration from unsupported
speculation.
\end{tcolorbox}

\vspace{1pt}
\begin{figure}[!tbp]
 \centering
 \includegraphics[width=1\linewidth]{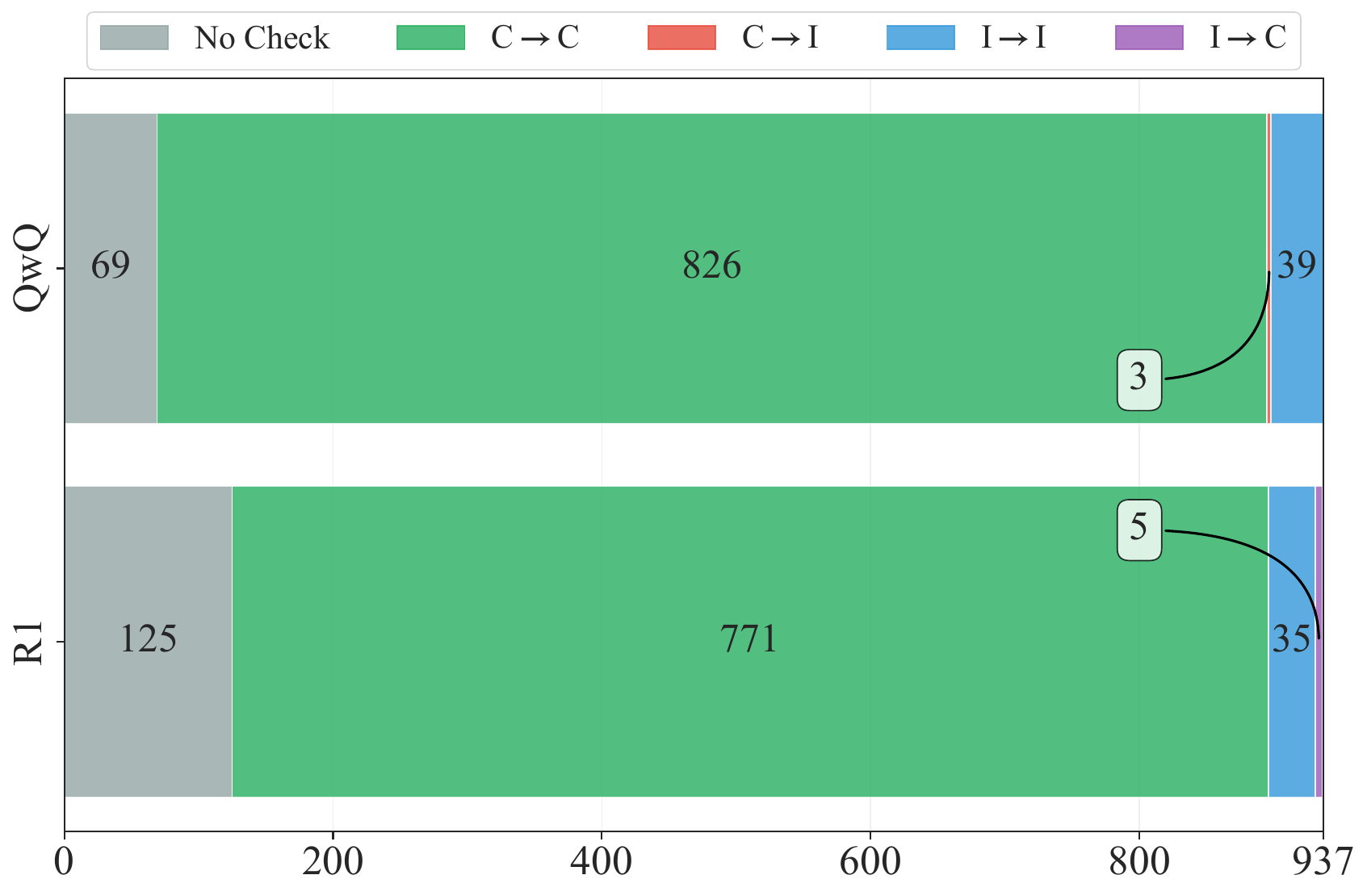}
 \caption{Post-answer check statistics. An A$\rightarrow$B transition denotes
 correctness before (A) and after (B) post-answer checking; for example,
 I$\rightarrow$C denotes an incorrect answer corrected to a correct one.}
 \label{fig:Post-Answer Check Distribution}
 \vspace{-1em}
\end{figure}

\textit{\textbf{Tedious reflections}.}
Reflection has been regarded as an important component of successful LRM
reasoning~\cite{DBLP:journals/corr/abs-2503-09567}. However, current LRMs
predominantly exhibit Reflection.Self-Monitoring Evaluation (R.SME), whereas
deeper reflective behaviours---such as identifying the cause of failure
through Reflection.Causal Attribution (R.CA) and subsequently adjusting the
strategy through Reflection.Strategy Regulation (R.SR) or
Suggestion.Branch Changing (S.BC)---remain much rarer.

A successful but uncommon example is shown below:

\begin{tcolorbox}[colframe=black]\small
\begin{enumerate}[leftmargin=*, nosep]
 \item[\textbf{160:}] \textbf{Reflection.Self-Monitoring Evaluation}:
 \textit{The problem says quadrilateral DEGF has area 288. But according to
 the coordinates given in Asymptote, the entire triangle has area 81, which
 contradicts the problem's statement that a part of it (DEGF) is 288.}
 \item[\textbf{161:}] \textbf{Reflection.Causal Attribution}:
 \textit{Wait, that's a problem. \textbf{The coordinates in the Asymptote
 code are just for drawing, but the actual problem has different scaling.}
 Therefore, using coordinates from Asymptote code directly would give wrong
 results. So maybe I shouldn't rely on those coordinates.}
 \item[\textbf{163:}] \textbf{Suggestion.Branch Changing}:
 \textit{Let me start over.}
\end{enumerate}
\end{tcolorbox}

Step 161 identifies the source of the contradiction detected in step 160 and
provides useful information for the subsequent change in strategy. In contrast,
many model trajectories restart reasoning without identifying why the previous
attempt failed.

We further analyse post-answer verification in two open-source LRMs,
DeepSeek-R1 and QwQ~\cite{qwq32b}. After producing an initial answer, these
models frequently generate an additional verification phase. However,
Figure~\ref{fig:Post-Answer Check Distribution} shows that such checks are
often superficial.

For DeepSeek-R1, only five failed answers are corrected after post-answer
checking, and all five changes arise from formatting corrections rather than
substantive logical revisions. QwQ fails to correct any of its incorrect
answers, while three initially correct instances are instead revised to
incorrect ones. In most cases, post-answer verification largely repeats the
previous reasoning with little new diagnostic information. Thus, the presence
of self-monitoring alone does not imply effective self-correction.

\paragraph{Targeted reflection intervention.}
To move beyond observational evidence, we additionally examine
\textbf{62 cases} in which DeepSeek-R1 or QwQ fails to correct an initially
incorrect answer. We truncate the trajectory after the first incorrect answer
and explicitly elicit richer reflection involving Causal Attribution (R.CA),
Counterfactual Thinking (R.CT), and Strategy Regulation (R.SR), using
corresponding instructions and in-context examples. Under this intervention,
\textbf{43 out of 62 cases (69.3\%)} are successfully corrected, while
19 remain incorrect.

This experiment does not isolate the causal contribution of each individual
reflection category, but it provides complementary intervention-based evidence
that explicitly eliciting richer reflection can improve failed
self-correction.

\begin{tcolorbox}[
title=\textbf{Takeaway 3},
colframe=white!90!black,
coltitle=black,
halign title=center,
outer arc=1mm]
Native post-answer reflection in current LRMs is predominantly shallow
self-monitoring and rarely produces substantive revisions. In our targeted
intervention, explicitly eliciting richer reflection processes corrects
\textbf{69.3\%} of previously failed self-correction cases, motivating future
methods that encourage comprehensive reflection rather than self-monitoring
alone.
\end{tcolorbox}

\begin{table}[tbp]
\centering
\begin{tabular}{c|ccc}
\toprule
\textbf{CoTs} & \textbf{Max} & \textbf{Min} & \textbf{Average} \\
\midrule
Before Intervention & 0.60 & 0.21 & 0.41 \\
After Intervention & \textbf{1.00} & \textbf{0.67} & \textbf{0.88} \\
\bottomrule
\end{tabular}
\caption{PNS ($\uparrow$) statistics before and after intervention.
Higher PNS indicates that the retained steps have greater estimated necessity
for the successful reasoning trajectory.}
\label{tab:pns-stats}
\vspace{-1em}
\end{table}

\vspace{1pt}
\textit{\textbf{Redundant thinking}.}
The preceding analyses suggest that many CoTs contain repeated or
non-essential steps, such as repetitive self-monitoring, that may contribute
little to the final answer. To quantify this redundancy, we adopt a
counterfactual intervention framework based on the Probability of Necessity
and Sufficiency (PNS)~\cite{yu2025causal}.

Rather than simply deleting a sentence, the intervention replaces a reasoning
step with an incompatible alternative and re-rolls the downstream reasoning
under the modified prefix. High PNS indicates that perturbing the step
consistently damages the original successful outcome, whereas low PNS suggests
that the step may be dispensable. We apply this analysis to ten representative
questions, with results summarised in Table~\ref{tab:pns-stats}; further
implementation details are given in Section~\ref{sec:pns_details}.

After intervention-based pruning, the average PNS increases from $0.41$ to
$0.88$, while the minimum PNS rises from $0.21$ to $0.67$. This indicates
that the original trajectories contain a substantial number of steps with low
estimated necessity.

\begin{tcolorbox}[
title=\textbf{Takeaway 4},
colframe=white!90!black,
coltitle=black,
halign title=center,
outer arc=1mm]
Our counterfactual analysis identifies substantial redundancy in current LRM
reasoning trajectories. Rather than simply increasing reasoning length, an
important direction is to develop objectives that encourage concise but
necessary reasoning and reduce computational overhead from dispensable steps.
\end{tcolorbox}

\paragraph{Robustness across models and domains.}
We additionally repeat our analysis on the newer MiniMax-M2 model and on
CodeForces reasoning traces. A.IO remains associated with successful reasoning
in both settings. On CodeForces, S.HG and S.AR also occur significantly later
in incorrect trajectories, whereas MiniMax-M2 attenuates this late-stage
pattern and exhibits more effective self-monitoring. These results suggest
that the taxonomy transfers across model generations and reasoning domains
while remaining sensitive to their behavioural differences. Full results,
together with segmentation, difficulty, and taxonomy-ablation analyses, are
reported in Appendix~\ref{sec:add_exp}.

\section{Conclusion}
Large reasoning models (LRMs) have demonstrated strong potential in web applications, yet their behaviours are difficult to interpret. Therefore, this paper presents a novel taxonomy for analysing reasoning behaviours in LRMs, establishing a bridge between computational methods and human cognitive processes. To enable scalable analysis, we propose CAPO, an automated framework that achieves expert-consistent labelling. Using CAPO in collaboration with domain experts, we construct a high-quality dataset of 277,534 labelled reasoning steps. Our empirical study on the dataset distils four key insights, offering actionable directions for future LRM improvements.

\section*{Limitations}
Despite the insights provided, this work contains several limitations. First, regarding data annotation, while our CAPO method serves as an effective auxiliary tool to augment data volume, the consistency between human and LLM annotators is not optimal; nevertheless, we verify that the core findings presented in Section~\ref{sec:exp_takeaways} remain consistent across both annotation subsets. Second, our analysis focuses on representative open-source LRMs (e.g., Deepseek R1 and Qwen QwQ) to ensure direct access to unaltered CoT outputs~\cite{ChenEtAl2025}, leaving the comparative analysis involving newer or closed-source models (e.g., Gemini~3) for future exploration. Finally, we primarily utilize mathematical problems to evaluate intrinsic reasoning due to the performance saturation on commonsense benchmarks, and we plan to extend our investigation to more complex real-world reasoning scenarios involving robustness and planning effectiveness~\cite{DBLP:journals/corr/abs-2502-09621} in future work.

\section*{Ethical considerations}
This work primarily utilizes publicly available mathematical reasoning datasets that do not contain personally identifiable information or offensive content. Regarding the human-annotated subset, the annotation process was conducted exclusively by the authors of this paper, ensuring domain expertise and adherence to data quality standards without involving external workers. We acknowledge that the LRMs analyzed herein may exhibit inherent hallucinations or biases. Consequently, our findings are intended to advance the scientific understanding of reasoning mechanisms and should not be interpreted as an endorsement for deploying these systems in high-stakes scenarios without further safety alignment.

\section*{Acknowledgement}
Lei Chen's work is partially supported by National Key Research and Development Program of China Grant No. 2023YFF0725100, National Science Foundation of China (NSFC) under Grant No. U22B2060, Guangdong-Hong Kong Technology Innovation Joint Funding Scheme Project No. 2024A0505040012, the Hong Kong RGC GRF Project 16213620, RIF Project R6020-19, AOE Project AoE/E-603/18, Theme-based project TRS T41-603/20R, CRF Project C2004-21G, Key Areas Special Project of Guangdong Provincial Universities 2024ZDZX1006, Guangdong Province Science and Technology Plan Project 2023A0505030011, Guangzhou municipality big data intelligence key lab, 2023A03J0012, Hong Kong ITC ITF grants MHX/078/21 and PRP/004/22FX, Hong Kong ITC TC-SKLCRCC26EG01, Zhujiang scholar program 2021JC02X170, Microsoft Research Asia Collaborative Research Grant, HKUST-Webank joint research lab, 2025 HKUST Shenzhen-Hong Kong Collaborative Innovation Institute Green Sustainability Special Fund from Shui On Xintiandi and the InnoSpace GBA, and HKUST(GZ) - CMCC(Guangzhou Branch) Metaverse Joint Innovation Lab under Grant No. P00659.

\bibliography{custom}

\appendix

\section{CoT Category Examples}\label{sec:appendix_cot}
Given below are explanations and real-world examples for each mental process selected from our annotated dataset. 

\subsection{Analysis}

\begin{cotdesc}
  \item[Category:] Analysis.Problem\_Definition
  \item[Explanation:] Identify and clearly define the core difficulty or central question in the task.
\end{cotdesc}
\begin{examplebox}
\textbf{Example:} ''Okay, let's see. I have this problem with triangle ABC. Points D and E are on AB, and F and G are on AC. The lengths are given: AD = 4, DE = 16, EB = 8, AF = 13, FG = 52, and GC = 26. The area of quadrilateral DEGF is 288. I need to find the area of the heptagon AFNBCEM.''
\end{examplebox}

\begin{cotdesc}
  \item[Category:] Analysis.Information\_Organization
  \item[Explanation:] List and organize all relevant background information and known facts.
\end{cotdesc}
\begin{examplebox}
\textbf{Example:}  \\
\textbf{Step 125:} ''Term1 = $0\times 45/7 - (24/7)\times 9 = -216/7$.  ''   \\ 
                    \textbf{Step 559:} ''Term1 = -216/7.''
\end{examplebox}

\begin{cotdesc}
  \item[Category:] Analysis.Problem\_Structuring
  \item[Explanation:] Decompose the main problem into smaller sub-problems and explain their logical relationships.
\end{cotdesc}
\begin{examplebox}
\textbf{Example:} ``Since D and E divide AB into 1:4:2, and F and G divide AC into 1:4:2, coordinates can be expressed in these ratios. Let me parametrize points D, E, F, G.''
\end{examplebox}

\subsection{Inference}

\begin{cotdesc}
  \item[Category:] Inference.Deductive\_Reasoning
  \item[Explanation:] Apply general principles or rules to deduce specific conclusions relevant to the current task.
\end{cotdesc}
\begin{examplebox}
\textbf{Example:} ``Area = $\tfrac{1}{2}\times|24|=12$. Therefore, the ratios on AB are $AD:DE:EB = 4:16:8$, so from A to D is 4/28 = 1/7, D to E is 16/28 = 4/7, and E to B is 8/28 = 2/7.''
\end{examplebox}

\begin{cotdesc}
  \item[Category:] Inference.Inductive\_Reasoning
  \item[Explanation:] Generalize patterns from specific examples or observations.
\end{cotdesc}
\begin{examplebox}
\textbf{Example:} ``Since $a=2b$ for $n=1, n=2, n=3$, then I suppose that $a=2b$ for any $n$.'' 
\end{examplebox}

\begin{cotdesc}
  \item[Category:] Inference.Abductive\_Reasoning
  \item[Explanation:] Given an observation, propose the most likely or plausible explanation.
\end{cotdesc}
\begin{examplebox}
\textbf{Example:} "Wait, maybe in the last period, the number of solutions is less, so the symmetry breaks. In previous periods, we had two solutions on each edge, so four solutions, and the sum cancels out. But in the last period, k=8, rising edge only has one solution and falling edge has one solution, so their sum is not zero."
\end{examplebox}

\subsection{Judgment}

\begin{cotdesc}
  \item[Category:] Judgment.Principle\_Selection
  \item[Explanation:] Choose appropriate logical principles or domain-specific rules needed to evaluate the problem.
\end{cotdesc}
\begin{examplebox}
\textbf{Example:} "3. **Derivative and Wilson's Theorem**:   - Using Wilson's theorem, $ P'(k) \equiv -1 \pmod{2027} )$.   - Therefore, $( \frac{-\sum\_{m \neq k} m}{P'(k)} \equiv k \pmod{2027} )$." 
\end{examplebox}

\begin{cotdesc}
  \item[Category:] Judgment.Evaluation\_of\_Alternatives
  \item[Explanation:] Compare multiple reasoning paths or hypotheses and select the most promising one.
\end{cotdesc}
\begin{examplebox}
\textbf{Example:} ``To count the number of solutions, perhaps we can use inclusion-exclusion, but that might be complex. Alternatively, since the grid is small, maybe we can compute it directly by degrees of freedom.''
\end{examplebox}

\begin{cotdesc}
  \item[Category:] Judgment.Conclusion\_Decision
  \item[Explanation:] Make a final decision or answer based on prior reasoning and comparisons.
\end{cotdesc}
\begin{examplebox}
\textbf{Example:} ``Therefore, triangle ABC has area 588.'' 
\end{examplebox}

\subsection{Suggestion}

\begin{cotdesc}
  \item[Category:] Suggestion.Strategic\_Planning
  \item[Explanation:] Develop a reasoning roadmap or outline for solving the problem.
\end{cotdesc}
\begin{examplebox}
\textbf{Example:} ``To compute the heptagon’s area, we can subtract the areas outside it from the area of triangle ABC.'' 
\end{examplebox}

\begin{cotdesc}
  \item[Category:] Suggestion.Branch\_Changing
  \item[Explanation:] Abandon the current reasoning path and explore a new or contrasting approach.
\end{cotdesc}
\begin{examplebox}
\textbf{Example:} ``Alternatively, use the shoelace formula with the coordinates.'' 
\end{examplebox}

\begin{cotdesc}
  \item[Category:] Suggestion.Hypothesis\_Generation
  \item[Explanation:] Generate a speculative explanation or assumption based on limited evidence.
\end{cotdesc}
\begin{examplebox}
\textbf{Example:} ``Perhaps $u$ is real, which could explain the rotated parabola’s points.'' 
\end{examplebox}

\begin{cotdesc}
  \item[Category:] Suggestion.Analogy\_Recall
  \item[Explanation:] Introduce an analogous situation or familiar pattern to guide reasoning.
\end{cotdesc}
\begin{examplebox}
\textbf{Example:} ``Since D is 1/7 along AB and F is 1/7 along AC, we can use the same ratio logic for E and G.'' 
\end{examplebox}

\subsection{Reflection}

\begin{cotdesc}
  \item[Category:] Reflection.Self\_Monitoring\_Evaluation
  \item[Explanation:] Review current reasoning steps for gaps, errors, or inconsistencies.
\end{cotdesc}
\begin{examplebox}
\textbf{Example:} ``Wait, if $q=42$, then area of ABC = 588. Shoelace formula gives DEGF = 288, which matches. So this is correct.''
\end{examplebox}

\begin{cotdesc}
  \item[Category:] Reflection.Counterfactual\_Thinking
  \item[Explanation:] Consider alternative actions and speculate on “what-if” scenarios.
\end{cotdesc}
\begin{examplebox}
\textbf{Example:} ``If we had divided by 22, the probability would be 0.045, but actual count shows 0.057 due to dependencies.'' 
\end{examplebox}

\begin{cotdesc}
  \item[Category:] Reflection.Causal\_Attribution
  \item[Explanation:] Analyze reasons behind success or failure by identifying key factors that caused the result.
\end{cotdesc}
\begin{examplebox}
\textbf{Example:} ``Because divisibility by 2 and 11 are not independent, our initial assumption failed.'' 
\end{examplebox}

\begin{cotdesc}
  \item[Category:] Reflection.Strategy\_Regulation
  \item[Explanation:] Adjust the overall problem-solving strategy based on reflection or feedback.
\end{cotdesc}
\begin{examplebox}
\textbf{Example:} ``Since I’m stuck, I should consult an official solution or try a different known formula.'' 
\end{examplebox}

\section{Discussion}\label{sec:discussion}
Furthermore, we will discuss the limitations and future work of the subsequent three aspects:

{\textbf{Suboptimal consistency between human and LLM as annotators}.} One may argue that although our CAPO effectively improves annotation quality, it still struggles to achieve a seemingly satisfactory level of consistency. Here, it is essential to emphasize that automated annotation serves only as an auxiliary method to augment the volume of human-labeled data. \textbf{We ensure that the findings presented above in Section~\ref{sec:exp_takeaways} remain consistent across both LLM-annotated and human-annotated subsets.}  For instance, in Takeaway 1, both annotation sets indicate that a higher A.IO ratio contributes to the model’s ability to answer questions correctly. The primary motivation for incorporating LLM-generated annotations in our study stems from the limited sample size in the human-annotated set, which undermines statistical reliability. Improving annotation accuracy, such as by incorporating more human-annotated data for post-training, represents an interesting direction for future work.

{\textbf{Reasoning models beyond Deepseek R1 and Qwen QwQ}.}
The primary motivations for selecting Qwen QwQ and Deepseek R1 as our subject models are twofold: (a) they represent leading open-source LRMs, enabling direct access to their unaltered CoT outputs; and (b) their core capabilities have been extensively validated in prior work~\cite{ChenEtAl2025}. Observing that different LRMs exhibit discernible variations in performance across different tasks or contexts, we propose that a comparative analysis involving newer opened models (e.g., updated version of R1) and representative closed-source LRMs with accessible APIs (e.g., Gemini~2.5) constitutes a significant avenue for future research.

{\textbf{Analysis of More complex real-world reasoning}.} As empirically observed during data collection, modern LRMs consistently achieve near-perfect accuracy on benchmark tasks assessing commonsense reasoning in daily-life scenarios (e.g., Commonsense QA). This high performance precludes the use of binary (correct/incorrect) evaluation, presenting significant challenges in assessing the quality of the generated reasoning chains. This difficulty extends to complex, real-world reasoning problems with greater severity. Consequently, we select mathematical problem-solving as our primary analytical focus. Mathematical problems offer sufficient complexity to thoroughly evaluate intrinsic reasoning capabilities while providing unambiguous response verification. In future work, we plan to examine how mental processes affect performance in complex reasoning scenarios from multifaceted perspectives, including robustness, computational efficiency, and planning effectiveness~\cite{DBLP:journals/corr/abs-2502-09621}.

\begin{figure*}[!t]
    \centering
    \includegraphics[width=0.8\linewidth]{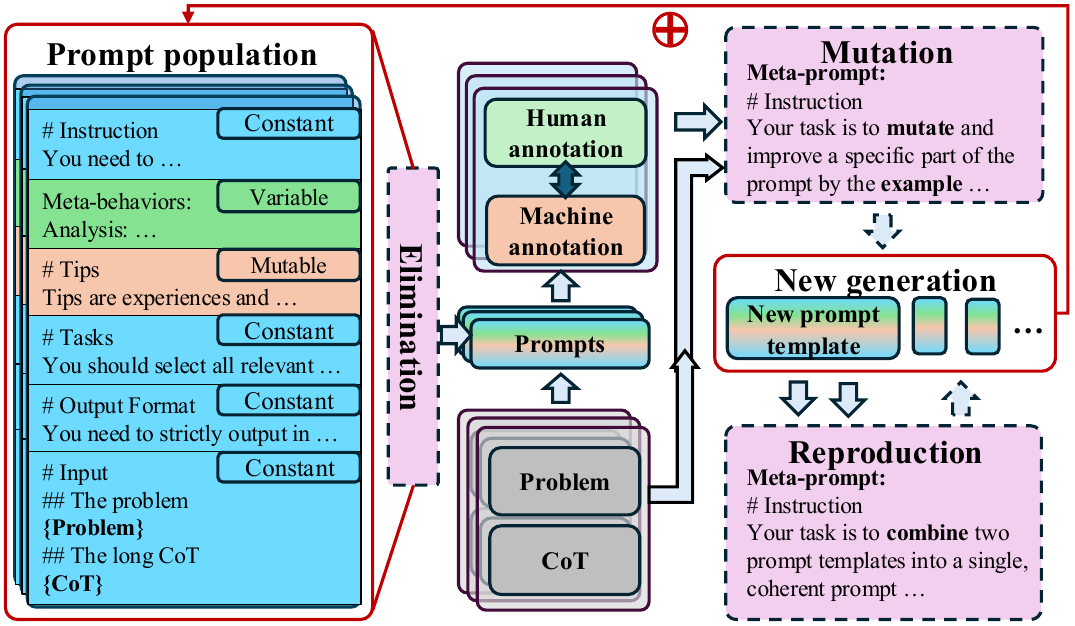}
    \caption{The CAPO framework, where each prompt comprises separated constant, variable, and mutable areas. }
    \label{fig:apo}
    
\end{figure*}

\section{Data Sources and Artifacts}\label{apx:data}
All four sources are available as AIME$^\star$\footnote{\url{https://huggingface.co/datasets/opencompass/AIME2025}}, HMMT\footnote{\url{https://huggingface.co/datasets/MathArena/hmmt\_feb\_2025}}, AIME\footnote{\url{https://huggingface.co/datasets/di-zhang-fdu/AIME\_1983\_2024}}, MATH\footnote{\url{https://huggingface.co/datasets/Maxwell-Jia/MATH}}. Beyond mathematics, we also annotated 1,030 common sense QA\footnote{\url{https://huggingface.co/datasets/peterkchung/commonsense\_cot\_partial\_raw}} CoTs, denoting as ComS and ComS$^\star$.

The source data utilised in this study are all publicly available and safe. Our annotation process did not involve any personally identifiable information (PII). Furthermore, the data focus is strictly on mathematics and commonsense QA, and we have verified through double-checks that no offensive content is present.

We confirm that our utilisation of existing datasets (e.g., AIME, MATH, HMMT) is strictly consistent with their intended use as academic benchmarks for evaluating reasoning capabilities. Regarding the new artifact introduced in this work (i.e., the fine-grained annotated dataset of atomic reasoning steps), its intended use is specified exclusively for research purposes, specifically for analysing and improving the interpretability of LRMs. This designation is fully compatible with the access conditions of the original source data, ensuring that the derivative artifacts are not employed for commercial or non-research applications in violation of the original licences.

\section{CAPO Implementation Details}\label{apx:capo_details}

\subsection{Algorithm and Settings}
The detailed workflow of CAPO is shown in Algorithm~\ref{alg:apo} and depicted in Figure~\ref{alg:apo}, where $n\_r, n\_m, n\_e, n\_0, g$ denote the numbers of reproduction, mutation, remaining candidates after elimination, initial mutations, and generations, respectively.

During evaluation, we use the following hyperparameters in Algorithm~\ref{alg:apo}: $n_r=4,n_m=5,n_e=8,n_0=10,g=4$ and $|G_{good}|=5$. The measurement function $M$ and elimination function $\phi$ are directly derived from the consistency metric. During optimization, the description of each meta-behavior (after ``\textit{meta behavior include:}'') is set as the variable area, and a new region named ``tips'' is the mutable area, while the remaining part is identified as the constant area. 

\begin{algorithm}[t] \small
    \caption{CAPO Algorithm}
    \begin{algorithmic}
        \REQUIRE Hyperparameters: $n_r,n_m,n_e,n_0,g>0$
        \REQUIRE Labeled long CoTs: $\{x_i,y_i\}_i$
        \REQUIRE Original population: $G\gets\{p_0\}$
        \REQUIRE Measurement function: $M(G)= G_\text{bad},G_\text{good}$
        \REQUIRE Mutation function: $\pi(p_\cdot,\{x,y\})=\tilde{p}$
        \REQUIRE Reproduction function: $\sigma(p_\cdot,p_\cdot)=\tilde{p}$
        \REQUIRE Elimination function: $\phi(G,n_e)=\tilde{G}$
        \WHILE{$n_0 > 0$}
            \STATE $n_0 \gets n_0-1$
            \STATE $\{x,y\} \gets \text{sample}(\{x_i,y_i\}_i)$
            \STATE $G \gets G\,\cup\,\{\pi(p_0,\{x,y\})\}$
        \ENDWHILE
        \WHILE{$g>0$}
            \STATE $g \gets g-1$
            \STATE $G_\text{bad},G_\text{good}\gets M(G)$
            \STATE $\{p_{i_k},p_{j_k}\}_{k=1}^{n_r} \gets \text{sample}_{n_r}(G_\text{good})$
            \FOR{$p_{i_k},p_{j_k}$}
                \STATE $G \gets G\,\cup\,\{\sigma(p_{i_k},p_{j_k})\}$ \\: \textbf{Reproduction}
            \ENDFOR
            \STATE $\{p_{k}\}_{k=1}^{n_m} \gets \text{sample}_{n_m}(G)$
            \STATE $\{x_k,y_k\}_{k=1}^{n_m} \gets \text{sample}_{n_m}(\{x_i,y_i\}_i)$
            \FOR{$p_k,\{x_k,y_k\}$}
                \STATE $G \gets G\,\cup\,\{\pi(p_{k},\{x_k,y_k\})\}$ \\: \textbf{Mutation}
            \ENDFOR
            \STATE $G\gets \phi(G,n_e)$ \\: \textbf{Elimination}
        \ENDWHILE
        \STATE \textbf{Return} $G$
    \end{algorithmic}
    \label{alg:apo}
\end{algorithm}

\subsection{Prompts}
The initial task prompt is as follows:
\begin{examplebox}
\small
\textbf{\# Instruction} \par
You need to classify meta-behaviors in an inputted chain of thought (CoT) when solving a problem. Each step is enclosed by <step *>, where * is the order number of a step. \par
Then, you should identify the meta-behaviors by each step. Details of the task are as follows. \par
\textbf{Meta-behaviors include:} \par
\textbf{\{Mental processes description\}} \par
\textbf{\# Task} \par
\textbf{\{Task description\}} \par
\textbf{\# Output Format}\par
\textbf{\{Output format control\}}
\textbf{\# Input}\par
\textbf{\#\# The problem}\par
\textbf{\{problem\_desc\}}\par
\textbf{\#\# The long CoT}\par
\textbf{\{CoT\}}\par
\textbf{\# Output}
\end{examplebox}

For the meta-prompts, the mutation prompt is as follows:
\begin{examplebox}
\small
\textbf{\# Instruction:}\par
You are an expert in prompt engineering.\par
The following five prompts describe a task.\par
You will also be given an example of a response to this prompt.\par
Your task is to mutate and improve a specific part of the prompt, based on the example and answer, according to the following rules:\par
(1) Review the task in the prompt to understand the key objectives and requirements that the instruction needs to address.\par
(2) Focus on the part of the prompt indicated by the <part> tag that needs mutation.\par
(3) Analyze the example and answer to identify any gaps, ambiguities or areas for improvement in the prompt.\par
(4) Maintain the original format and structure of the prompt while enhancing clarity, specificity and guidance in the mutated part.\par
(5) Ensure that the output format of the mutated part is consistent with the original prompt structure.\par
(6) Do not modify the names of meta-behaviors or the structure of the prompt.\par
(7) Tags (e.g., in <>) should not be modified.\par
(8) All names of meta-behaviors (xx.xx) should be strictly retained as they are, except for their descriptions, which can be modified for clarity. And, no addition and deletion of meta-behaviors is allowed.\par
(9) Particularly, some absent fields in the example typically indicate that the original response is with incorrect format, try to fix it.\par
(10) Especially, tips, including details of meta-behaviors and tasks, are more flexible and can be modified to better fit the merged prompt.\par
(11) Notice that, example is not available when the following prompt is used for downstream tasks.\par

Output the specific mutated part in the <mutated\_part> tag after providing a comprehensive, step-by-step thinking, as follows:\par
**Output format: <mutated\_part> Mutated and improved part of the prompt </mutated\_part>**\par

\{\textbf{Current prompt}\}

\textbf{\# Example}\par
\textbf{\{example\}}\par

\textbf{\# Mutation Target}\par
\textbf{\{part\_name\}}\par
\end{examplebox}

And, the reproducibility meta-prompt is:
\begin{examplebox}
\small
\textbf{\# Instruction}\par
You are an expert in prompt engineering.\par
You will be given two prompts, each consisting of five parts that describe a task.\par
Each prompt is optimised based on an example of the task.\par
Your task is to combine these two prompts to create a single, coherent prompt that retains the strengths of both while ensuring clarity and consistency.\par
Focus on the following aspects:\par
(1) Identify the key objectives and requirements in both prompts.\par
(2) Combine the relevant parts from both prompts to create a comprehensive and clear merged prompt.\par
(3) Maintain the original format and structure of the prompts, enhancing clarity, specificity and guidance where necessary.\par
(4) Ensure the merged prompt is logically consistent and flows well.\par
(5) The output format should be consistent with the original prompt structure.\par
(6) Do not modify the names of meta-behaviors or the structure of the prompt.\par
(7) Tags (e.g., in <>) should not be modified.\par
(8) All names of meta-behaviors (xx.xx) should be strictly retained as they are, except for their descriptions, which can be modified for clarity. And, no addition and deletion of meta-behaviors is allowed.\par
(9) The consideration aspects of both prompts may vary, so you should carefully merge them to ensure that the final prompt is comprehensive.\par
(10) Especially, tips, including details of meta-behaviors and tasks, are more flexible and can be modified to better fit the merged prompt.\par

Output the merged prompt with the five tags after providing comprehensive step-by-step reasoning as follows:\par
\textbf{\textit{Output format omitted}}\par

\textbf{\# Prompt 1}\par
\{\textbf{\textit{Current prompt 1}}\}

\textbf{\# Prompt 2}\par
\{\textbf{\textit{Current prompt 2}}\}
\end{examplebox}

Please refer to our source code for concrete prompts.

\subsection{Experimental Setup}
\paragraph{Model Selection.}
We employ \textit{Gemini-2.5-flash-preview-05-20} (configured without the ``thinking mode'') as the backbone LLM annotator, as preliminary experiments indicated it offered the best balance of performance and cost among accessible APIs.

\paragraph{Dataset and Metric.}
To evaluate annotation quality, we utilize the human-annotated CoTs from the AIME$^\star$ and HMMT datasets. We randomly partition these into a training set ($50\%$) for the CAPO optimization process and a test set ($50\%$) for validation. The primary metric, \textbf{consistency}, is defined as the average proportion of reasoning steps where the LLM's assigned category strictly matches the human expert's annotation.

\subsection{Baseline Implementation (RAG)}
We implement a Retrieval-Augmented Generation (RAG) baseline to benchmark the effectiveness of ICL for this task. The workflow is as follows:
\begin{enumerate}
    \item \textbf{Embedding:} We use \textit{Linq-Mistral}~\cite{LinqAIResearch2024} to encode all CoTs in the training set into vector representations.
    \item \textbf{Retrieval:} For each query CoT in the test set, we calculate the inner product similarity to retrieve the most semantically similar example from the training set.
    \item \textbf{Prompt Construction:} The retrieved example, along with its ground-truth human annotations and paired zero-shot predictions, is injected into the prompt context. This allows the LLM to perform few-shot learning by observing how human experts correct zero-shot errors.
\end{enumerate}
Despite utilising retrieval to optimise context relevance, the RAG baseline is constrained by the token limit and fails to outperform the optimised instructions generated by CAPO.

\section{Details of Human Annotators}\label{sec:human_annotators}

The human annotation process was conducted exclusively by the authors of this paper to ensure a high level of domain expertise and consistency. Consequently, all annotators were fully aware of and consented to the intended use of the data. Due to the anonymity policy, we are unable to provide detailed profiles of the annotators. However, we ensure that all participants involved either hold or are currently pursuing advanced degrees (e.g., PhD), and the annotation and analysis workflow was led by winners of mathematics competitions. 

Prior to the formal annotation, all annotators underwent multiple rounds of training to familiarise themselves with the taxonomy proposed in Section~\ref{sec:Human-Thought-Taxonomy}. We conducted trial annotations on a pilot subset of the data, where any discrepancies were analysed and resolved through subsequent discussions. This iterative process continued until all annotators achieved a high level of consensus and fully mastered the annotation criteria.\par

To alleviate the manual workload and improve efficiency, we adopted a human-in-the-loop strategy. We first utilised the proposed CAPO framework to pre-generate initial annotations for the dataset. Human annotators were then tasked with reviewing these pre-populated labels and performing necessary corrections. This approach allowed annotators to focus their attention on verifying complex reasoning steps rather than starting from scratch.

\section{Additional Experiments and Robustness Analyses}
\label{sec:add_exp}

We conduct additional experiments to examine whether the observations in
Section~\ref{sec:exp_takeaways} generalise across newer reasoning models,
different reasoning domains, alternative segmentation strategies, and
different levels of taxonomic granularity.

\subsection{Generalisation to a Newer LRM}

We first extend our analysis to MiniMax-M2, a more recent reasoning model.
Using the same taxonomy and CAPO annotation pipeline, we analyse reasoning
trajectories generated on the mathematical datasets used in our main
experiments. Table~\ref{tab:minimax_stats} summarises the resulting data.

\begin{table}[!tbp]
\centering
\begin{tabular}{c|cc}
\toprule
\textbf{Correctness} & \textbf{CoTs} & \textbf{Steps} \\
\midrule
Correct   & 1,803 & 740,399 \\
Incorrect & 118   & 87,965  \\
\bottomrule
\end{tabular}
\caption{Statistics of the MiniMax-M2 reasoning trajectories.}
\label{tab:minimax_stats}
\vspace{-1em}
\end{table}

MiniMax-M2 achieves an accuracy of \textbf{93.85\%} in this setting, compared
with \textbf{86.63\%} for DeepSeek-R1, and produces substantially longer
reasoning trajectories on average (\textbf{431.21} versus \textbf{138.62}
steps).

Despite these differences, \textbf{A.IO remains a significant differentiator}
between successful and unsuccessful reasoning, supporting the robustness of
Takeaway~1. At the same time, the behaviour of the newer model provides useful
nuance to the other findings. In particular, the relative positions of S.HG
and S.AR no longer differ significantly between correct and incorrect
trajectories ($p>0.9$ for both), suggesting that MiniMax-M2 substantially
attenuates the late-stage speculative pattern observed in DeepSeek-R1.

We also observe that R.SME is positively associated with correct answers in
MiniMax-M2, unlike in the earlier models. Nevertheless, its reflective
behaviour remains dominated by self-monitoring rather than deeper processes
such as R.CA. Thus, the taxonomy captures not only patterns that persist across
model generations, but also behavioural changes in newer LRMs.

\subsection{Generalisation to Code Reasoning}

To examine cross-domain generalisation, we further apply our framework to
reasoning trajectories collected from the \texttt{open-rl/codeforces}
dataset. The resulting dataset contains 966 CoTs and over 420K annotated
reasoning steps, as shown in Table~\ref{tab:codeforces_stats}.

\begin{table}[!tbp]
\centering
\begin{tabular}{c|cc}
\toprule
\textbf{Correctness} & \textbf{CoTs} & \textbf{Steps} \\
\midrule
Correct   & 644 & 225,761 \\
Incorrect & 322 & 195,396 \\
\bottomrule
\end{tabular}
\caption{Statistics of the CodeForces reasoning trajectories.}
\label{tab:codeforces_stats}
\vspace{-1em}
\end{table}

Several principal patterns observed in mathematical reasoning also appear in
code reasoning. First, A.IO remains significantly more prevalent in successful
trajectories, with a mean difference of \textbf{$+0.03$} ($p<0.01$).

Second, S.HG and S.AR again occur substantially later in incorrect
trajectories. Table~\ref{tab:codeforces_position} shows that the average
relative position of S.HG shifts from $0.239$ in correct solutions to $0.346$
in incorrect ones, while S.AR shifts from $0.298$ to $0.410$; both
differences are significant at $p<10^{-6}$.

\begin{table}[!tbp]
\centering
\begin{tabular}{c|ccc}
\toprule
\textbf{Name} & \textbf{Pos. co.} & \textbf{Pos. inco.}
& \textbf{P-value} \\
\midrule
S.HG & 0.239 & 0.346 & $<10^{-6}$ \\
S.AR & 0.298 & 0.410 & $<10^{-6}$ \\
\bottomrule
\end{tabular}
\caption{Relative positions of S.HG and S.AR on CodeForces.}
\label{tab:codeforces_position}
\vspace{-1em}
\end{table}

Finally, reflective categories do not occur at larger proportions in
successful CodeForces trajectories, and deeper reflection remains uncommon.
Together, these results suggest that the principal patterns underlying
Takeaways~1--3 are not restricted to mathematical problem solving.

For commonsense reasoning, DeepSeek-R1 achieves near-perfect performance,
making correct--incorrect comparisons less informative. Qualitative inspection
instead reveals frequent stereotypical self-monitoring (R.SME) with little
subsequent causal attribution (R.CA), consistent with the distinction between
surface monitoring and deeper reflection identified in Takeaway~3.

\subsection{Effects of Difficulty and Reasoning Length}

A potential concern is that the observed mental-process patterns could simply
reflect problem difficulty or CoT length. We therefore analyse the association
between process density, MATH difficulty level, and reasoning length.

The resulting correlations show that S.HG increases as problems become more
difficult and reasoning chains grow longer, consistent with the observation
that hypothesis generation becomes more common when reasoning requires
extended exploration. A.IO also exhibits positive correlations with both
difficulty and CoT length, indicating that information organisation becomes
increasingly relevant as trajectories grow more complex. The analysis further
shows a negative correlation of S.AR density with both difficulty and CoT
length. Together with its temporal analysis in Section~\ref{sec:exp_takeaways},
these results indicate that the \emph{position} and \emph{context} in which
these processes occur are important, rather than their mere presence.

\subsection{Robustness to Step Segmentation}

Our default segmentation uses the natural
`\textbackslash{}n\textbackslash{}n' boundaries generated by LRMs.
Across the full annotated corpus, \textbf{82.2\%} of paragraph-level segments
contain exactly one mental-process category, approximately \textbf{16\%}
contain two, and only \textbf{2\%} contain more than two. This indicates that
the natural paragraph boundaries usually correspond to semantically coherent
reasoning units.

We further perform a robustness analysis on 100 randomly sampled CoTs using
sentence-level splitting. On this subset, sentence-level segmentation
increases the fraction of single-category segments from \textbf{73.2\%} to
\textbf{89.0\%}, but also increases the average number of annotation units
from \textbf{111.7} to \textbf{182.3} per CoT, corresponding to a
\textbf{1.63$\times$} increase in annotation cost.

The overall mental-process distributions remain broadly consistent between
the two segmentation strategies. The largest shifts are an increase of
\textbf{6.15\%} in I.DR and a decrease of \textbf{3.34\%} in A.IO under
sentence-level segmentation, which is expected because long derivations are
split into individual sentences while information organisation often operates
at the paragraph level. These results support paragraph-level segmentation as
a practical balance between semantic coherence, analytical granularity, and
annotation cost.

\subsection{Fine- versus Coarse-Grained Taxonomy}

To examine whether the seventeen fine-grained processes provide information
beyond the five parent categories, we aggregate the labels into
\textit{Analysis}, \textit{Inference}, \textit{Judgment},
\textit{Suggestion}, and \textit{Reflection}, and repeat the
correct--incorrect comparison.

\begin{table}[!tbp]
\centering
\resizebox{0.92\columnwidth}{!}{\begin{tabular}{c|ccc|c}
\toprule
\textbf{Category} & \textbf{Correct} & \textbf{Incorrect}
& \textbf{Diff.} & \textbf{Sig.} \\
\midrule
Analysis    & 0.2480 & 0.2117 & $+0.0363$ & Yes \\
Inference   & 0.6291 & 0.6467 & $-0.0176$ & No  \\
Judgment    & 0.1372 & 0.0799 & $+0.0573$ & Yes \\
Suggestion  & 0.1474 & 0.1507 & $-0.0033$ & No  \\
Reflection  & 0.2330 & 0.2214 & $+0.0116$ & No  \\
\bottomrule
\end{tabular}}
\caption{Correct--incorrect comparison after aggregating the seventeen
fine-grained processes into five parent categories.}
\label{tab:coarse_ablation}
\vspace{-1em}
\end{table}

As shown in Table~\ref{tab:coarse_ablation}, coarse categories can obscure
process-specific signals. For example, the aggregate Suggestion category shows
almost no overall difference between correct and incorrect CoTs
($-0.0033$), even though its S.HG and S.AR subcategories exhibit clear
temporal patterns associated with unsuccessful reasoning. Similarly, the
aggregate Analysis category cannot distinguish the specific contribution of
A.IO from the other analysis processes.

This ablation does not establish that our seventeen-category taxonomy is the
unique decomposition of LRM reasoning. It instead demonstrates that the
fine-grained representation preserves behavioural information that can be
diluted when heterogeneous processes are merged.

\subsection{Counterfactual PNS Intervention}
\label{sec:pns_details}

Finally, we clarify the counterfactual analysis used in Takeaway~4.
Our PNS intervention is not a simple deletion test. Given an original
reasoning trajectory, we replace a target step with a corrupted or incompatible
alternative and then re-roll the downstream trajectory under this modified
prefix. This downstream-adaptive continuation allows subsequent reasoning to
respond naturally to the intervention rather than forcing the remainder of
the original CoT to remain fixed.

For each target step, PNS is estimated from repeated counterfactual rollouts
using final-answer correctness as the outcome evaluator. Intuitively, if
perturbing a step repeatedly causes an originally successful trajectory to
fail, the step receives a high PNS and is regarded as important to the
successful reasoning process. Conversely, a low PNS indicates that the
trajectory can often recover from perturbing that step, suggesting that the
step may be redundant or dispensable.

We apply this analysis to ten representative questions. As reported in
Table~\ref{tab:pns-stats}, intervention-based pruning increases the average
PNS of retained reasoning from $0.41$ to $0.88$, while the minimum increases
from $0.21$ to $0.67$. We therefore interpret the result as evidence of
substantial step-level redundancy in these analysed trajectories, while noting
that this counterfactual experiment is conducted on a relatively small subset.

\balance
\end{document}